\documentclass[conference]{IEEEtran}

\usepackage{color}
\usepackage{graphicx}
\usepackage{cite}
\usepackage{multirow}
\usepackage{url}
\usepackage{diagbox}
\usepackage{algorithm}
\usepackage{algorithmicx}
\usepackage{algpseudocode}
\usepackage{amsmath, bm}
\usepackage{hyperref}
\usepackage{booktabs}
\usepackage[OT1]{fontenc} 
\begin{document}
\title{DMOFC: Discrimination Metric-Optimized \\Feature Compression \vspace{-1em}}
\author{\IEEEauthorblockN{Changsheng Gao, Yiheng Jiang, Li Li, Dong Liu, and Feng Wu}
		University of Science and Technology of China, Hefei, China\\
		\{changshenggao, lil1, dongeliu, fengwu\}@ustc.edu.cn;
		bitaswood@mail.ustc.edu.cn}

\maketitle

\begin{abstract}
%\boldmath

Feature compression, as an important branch of video coding for machines (VCM), has attracted significant attention and exploration. However, the existing methods mainly focus on intra-feature similarity, such as the Mean Squared Error (MSE) between the reconstructed and original features, while neglecting the importance of inter-feature relationships. In this paper, we analyze the inter-feature relationships, focusing on feature discriminability in machine vision and underscoring its significance in feature compression. To maintain the feature discriminability of reconstructed features, we introduce a discrimination metric for feature compression. The discrimination metric is designed to ensure that the distance between features of the same category is smaller than the distance between features of different categories. Furthermore, we explore the relationship between the discrimination metric and the discriminability of the original features. Experimental results confirm the effectiveness of the proposed discrimination metric and reveal there exists a trade-off between the discrimination metric and the discriminability of the original features.
\end{abstract}
\IEEEpeerreviewmaketitle

\section{Introduction}
The exponential growth of image and video data has rendered it impractical for humans to manually analyze all of them. With the remarkable performance of machine vision algorithms, the analysis of image and video data has shifted towards machines by leveraging the capabilities of vision algorithms. 
Consequently, an increasing volume of image and video content is being processed by machines for analysis. 
However, prior to any analysis, it is crucial to store and transmit these data efficiently, leading to the emergence of the research topic known as video coding for machines (VCM).

Feature compression, as an important branch in VCM, has attracted considerable attention and exploration. Based on distortion measurement methods, feature encoding can be broadly categorized into three types.
The first kind of method, similar to image compression, measures distortion by the Mean Squared Error (MSE) between the reconstructed feature and the original feature \cite{choi2018deep, suzuki2022deep, kim2023end, liu2023learnt}. For example, in \cite{kim2023end}, the feature MSE is used for multi-layer feature compression.
In the second kind of method, researchers incorporate semantic information into the distortion measurement \cite{choi2021latent, chen2019lossy, chen2020toward, ikusan2021rate, hu2020sensitivity}.
Departing from MSE constraints between reconstructed and original features, this method opts for constraining the semantic similarity between reconstructed and original features. This approach provides a more direct way of constraining the semantic information within the reconstructed features. For instance, in \cite{chen2019lossy}, one-hot vectors of the original and reconstructed features are obtained, and the similarity between them serves as the distortion measure.
In the third kind of method, researchers adopt a more direct semantic information constraint to reconstructed features \cite{singh2020end, yang2021video, zhang2021MSFC, yan2021SSSIC, alvar2019multi}. In this approach, they directly employ the loss function of visual tasks as the distortion measure for feature compression. For example, the machine vision algorithm is treated as the distortion of the compression network.

Existing methods adhere to the principles of image compression, where each individual feature undergoes independent optimization during the encoding process. For instance, the similarity measure MSE is computed on a single feature and is unrelated to the others. Despite the substantial progress achieved by these methods, this independent compression approach neglects the relationships between different features. However, for features utilized in machine vision, the relationships between different features, such as similarity and discriminability, play a pivotal role in the final machine vision analysis. Current independent compression methods fail to consider these inter-feature relationships, limiting the efficiency of feature compression.

In this paper, we investigate the inter-feature relationship in the context of the discriminability of features. Discriminability is one of the most important characteristics of features because it is the basis of the classification. Given classification plays an important role in various machine vision tasks, feature discriminability becomes pivotal for machine vision tasks. Our contributions, derived from a thorough investigation of feature discriminability, can be summarized as follows:
\begin{itemize}
	\item We are the first to address the importance of inter-feature relationships in feature compression and study it in the context of feature discriminability
	\item We propose a discrimination metric-optimized feature compression and verified its effectiveness for feature compression in person re-identification (person ReID) and face verification (FV) task 
	\item We study the influence of the discriminability of original features on the discrimination metric, and unveil the trade-off between them
\end{itemize}

\begin{figure}[tbp]
    \centering
    \begin{minipage}{\linewidth}
        \includegraphics[width=\linewidth]{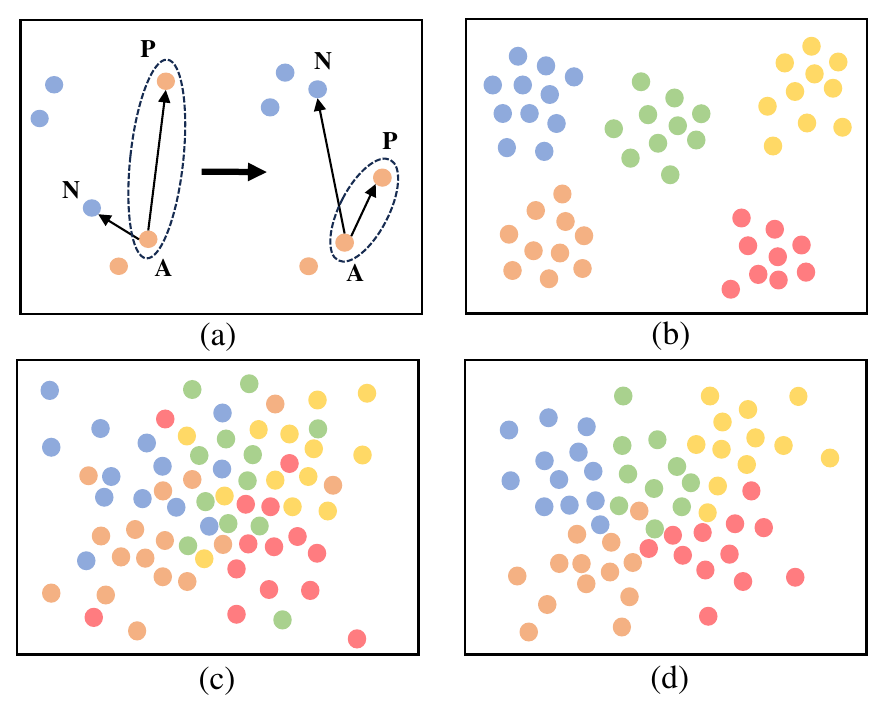}
    \end{minipage}
    \caption{Illustration of the discrimination metric and its comparison with the similarity metric. (a) The illustration of the discrimination metric. (b) An exemplary visualization of original features. (c) An exemplary visualization of reconstructed features optimized by the similarity metric. (d) An exemplary visualization of reconstructed features optimized by the discrimination metric.}
    \label{fig:metric}
\end{figure}

\section{Method}
In this section, we first analyze and compare the quality assessment of image and feature compression. Then, we introduce the proposed discrimination metric. After that, the specific feature compression method is presented. Finally, we introduce the extraction of the original feature and optimization strategy.
\subsection{Quality Assessment Analyses}
In this subsection, we compare the quality assessment of images for human vision and features for machine vision.

Since natural images are captured for human vision, the image quality assessment is conducted by human eyes. When a human assesses the quality of an image, the evaluation is performed on a single image, and no other images are involved. 
Given this independent quality assessment process, the distortion measure for image compression can also be independent, which means the image distortion measure can be performed on a single image. For example, based on the belief that higher similarity between the reconstructed and original image leads to higher quality, similarity metrics such as MSE have been widely employed in image compression methods.

However, the logic of quality assessment changes in the feature domain. The inter-feature relationship is usually considered in machine vision algorithms. Therefore, evaluating the quality of a specific feature during compression without considering its relationship to others would be inappropriate.
Taking the most fundamental task, image classification, as an example, feature quality assessment includes at least two aspects. First, it is assessed by measuring how similar the reconstructed feature is to the original feature. Second, it is assessed by measuring how discriminative a specific reconstructed feature is compared to other reconstructed features.
However, existing feature compression methods only consider similarity. Without discriminability considered, it can be imagined that at low bitrates, the reconstructed features would deviate from the original features, leading to a degradation of feature discriminability and, consequently, worse machine vision performance.

\subsection{Discrimination Metric}
With the above analyses, we believe the consideration of feature discriminability can improve the ultimate machine vision performance.

Maintaining inter distance between features of different categories is crucial for preserving discriminability among them. To maintain the inter distance, we propose a discrimination metric that consists of three samples: an anchor (A), a positive (P), and a negative (N). The anchor and positive belong to the same category, while the negative belongs to a different category. By constraining the positive to be close to the anchor and the negative to be distant from the anchor, we enhance the discriminability between different categories.
The discrimination metric is computed as: 
\begin{equation}
    DIS(A,P,N) = \max(d(A,P)-d(A,N)+\alpha,0)
    \label{equ:DIS}
\end{equation}
where $d$ denotes the distance measure MSE, and $\alpha$ is the margin that defines the minimum difference between the distance of A to P and the distance of A to N. $\alpha$ is set to 0.3 in our method. The mechanism of the discrimination metric is illustrated in Fig. \ref{fig:metric}
(a), where the distance of features from different classes is constrained larger than the distance of features from the same class. 

For each optimization iteration, the discrimination metric is applied to a few features. By iterating through all the features in the training set, we can preserve the discriminability of all features. We illustrate a schematic comparison between the similarity metric and discrimination metric in Fig. \ref{fig:metric} (c) and (d). The similarity metric tends to result in overlap as it fails to constrain the inter-class distance between features of different categories. In contrast, by incorporating inter-class distance constraints, the discrimination metric alleviates the phenomenon of overlap.
\subsection{Feature Compression}
Given that the extracted features are represented as 1D vectors ($1\times2048$ for person ReID and $1\times1024$ for face verification), we employ a simple feature compression method consisting of two fully-connected (FC) layers. The first FC layer serves as the encoder, while the other acts as the decoder. 

To comprehensively assess the impact of the discrimination metric, we vary the information capacity (IC) constraints by adjusting the output channels of the encoder and the input channels of the decoder. Specifically, we set five information capacities: 16, 32, 64, 128, and 256 bytes. 

\subsection{Optimization Strategy}
The compression network is optimized by the features extracted from the machine vision tasks. For the person ReID task, we extracted the features from the last layer of the FastReID algorithm \cite{he2020fastreid} implemented with a ResNet50 architecture. For the face verification task, 
the features are extracted from the last layer of the ResNet18-based ArcFace algorithm \cite{deng2019arcface}.

During training, for each optimization iteration, we randomly select 8 categories, and for each category, 16 samples are chosen randomly. Within each category, we calculate its discrimination distance with other categories using Equation \ref{equ:DIS} for all samples. Then, we select the sample with the largest discrimination distance to perform backpropagation. This approach allows the compression network to focus on the most challenging cases, aiding in effective network optimization.
\section{Experiments}
\begin{figure*}[tbp]
	\centering
	\begin{minipage}{0.4\linewidth}
		\includegraphics[width=\linewidth]{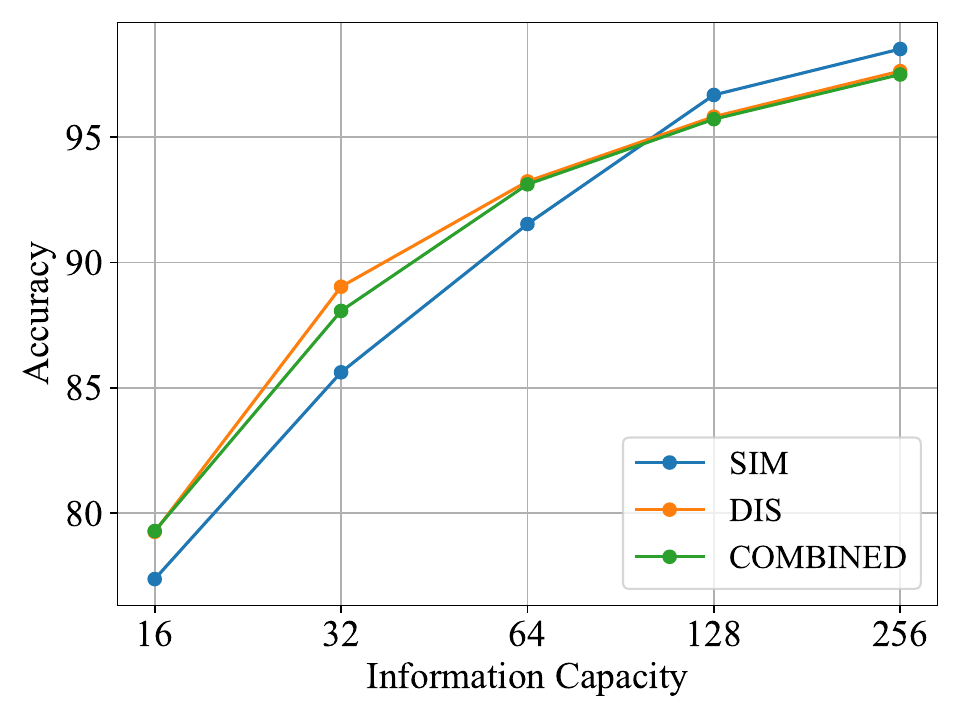}
		\centerline{\footnotesize{(a)}}
	\end{minipage}
	\begin{minipage}{0.4\linewidth}
		\includegraphics[width=\linewidth]{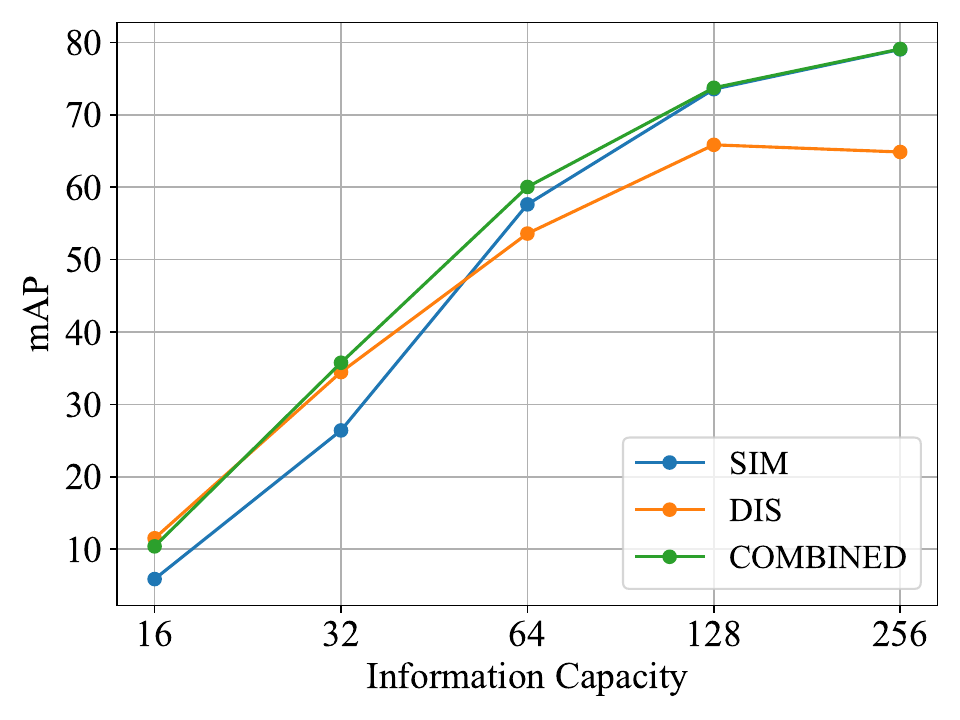}
		\centerline{\footnotesize{(b)}}
	\end{minipage}
	\caption{Evaluation performance comparisons between different metrics. (a) Face verification task. (b) Person re-identification task. }
	\label{fig:main_performance}
\end{figure*}
\subsection{Experimental Settings}
In this subsection, we introduce dataset and training configurations.
\subsubsection{Dataset} 
For the face verification task, we use features extracted from the CASIA-WebFace dataset \cite{yi2014learning} for training and evaluate the model on the Labeled Faces in the Wild (LFW) dataset \cite{LFWTech}. 
For the person ReID task, we use features extracted from the DukeMTMC-reID dataset \cite{zheng2017unlabeled}. 

\subsubsection{Training Configuration} 
For the face verification task, we randomly select 8 classes with 16 samples each for one iteration. We train for 300 epochs with 3125 iterations per epoch.
For the person ReID task, we utilize the sbs\_R50\_ibn configuration, setting the batch size and number of epochs to 512 and 200, respectively. 
For both tasks, the initial learning rate is set to 0.001 and scheduled with the cosine annealing algorithm. The Adam optimizer \cite{kingma2014adam} is employed for gradient optimization.
\subsection{Performance Analyses}
In this subsection, we compare the evaluation performance of different metrics on the two tasks. Specifically, SIM and DIS denote the similarity metric MSE and discrimination metric in Equation \ref{equ:DIS}, respectively. COMBINED denotes the combination of them. The combination of the two metrics is weighted equally in our experiments.
\subsubsection{Face Verification} 
The evaluation performance of three metrics on the face verification task is presented in Fig. \ref{fig:main_performance} (a). Generally, DIS metric is comparable with COMBINED metric. At low information capacities, both DIS and COMBINED outperform SIM. However, at high information capacities, SIM achieves higher evaluation performance than both discrimination metrics. These results demonstrate that discrimination metrics work well at low information capacities but fail at high information capacities. As we know, with sufficient information capacity, the signal fidelity of the original features can be well maintained. Given that the original features are discriminative and compact, once the signal fidelity is guaranteed, high evaluation performance can be achieved. From the curves in Fig. \ref{fig:main_performance} (a), the trade-off point of discrimination metrics and similarity metrics for the face verification task is located in an information capacity between 64 bytes and 128 bytes.
\subsubsection{Person ReID} 
The evaluation performance of three metrics on the person ReID task is presented in Fig. \ref{fig:main_performance} (b). DIS outperforms SIM at low information capacities but achieves lower performance at high information capacities. At low information capacities, achieving high signal fidelity is challenging, resulting in low discriminability between features. In this scenario, DIS helps improve the discriminability between features, leading to a performance improvement. However, at high information capacities, signal fidelity can be well maintained by optimizing the similarity between original features and reconstructed features through SIM. In this case, building better discriminability from scratch using DIS becomes challenging compared to preserving discriminability using SIM.
COMBINED metric achieves higher performance than SIM at all information capacities. On one hand, at low information capacities, COMBINED metric leverages DIS to reconstruct more discriminative features. On the other hand, at high information capacities, COMBINED utilizes SIM to preserve the discriminability between the original features.

\subsection{Feature Analyses} 
\begin{figure*}[tbp]
	\centering
	\begin{minipage}{0.32\linewidth}
		\includegraphics[width=\linewidth]{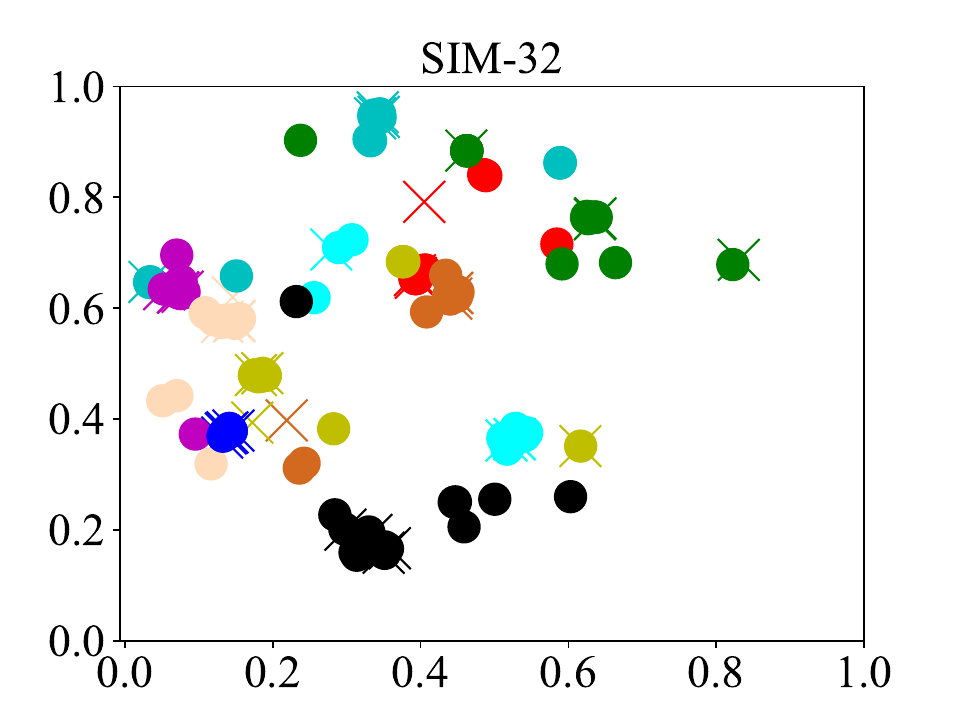}
	\end{minipage}
	\begin{minipage}{0.32\linewidth}
		\includegraphics[width=\linewidth]{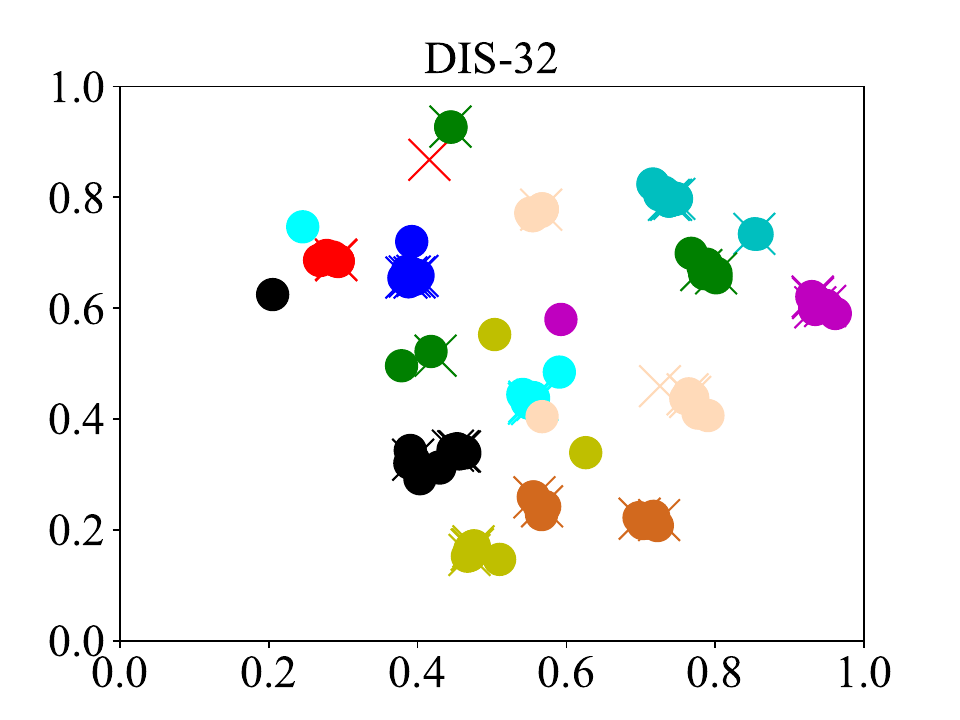}
	\end{minipage}
	\begin{minipage}{0.32\linewidth}
		\includegraphics[width=\linewidth]{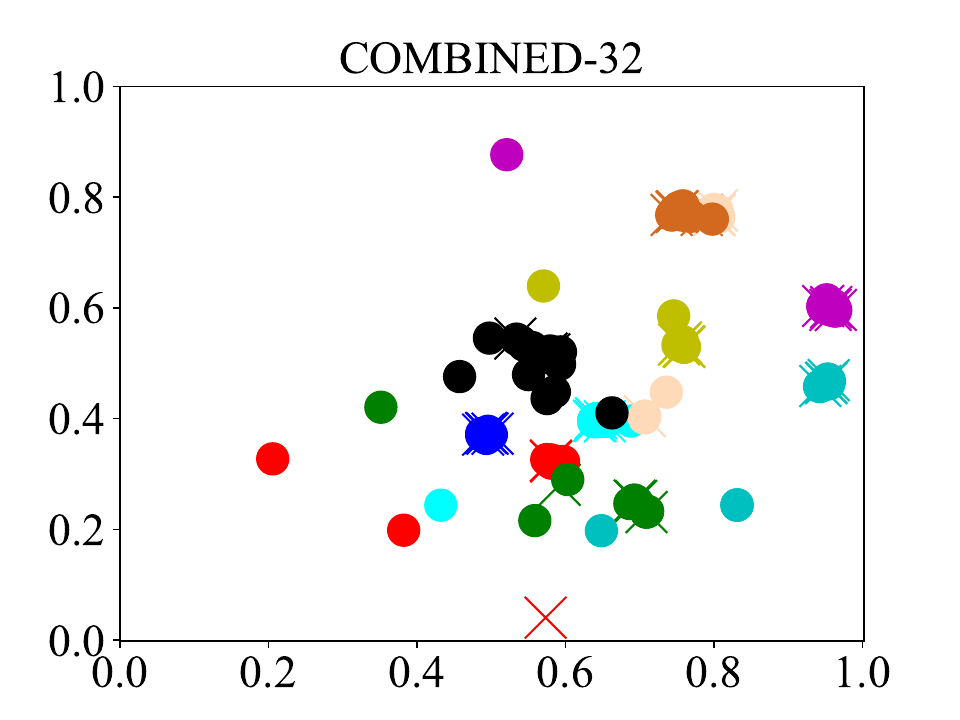}
	\end{minipage}
	
	\begin{minipage}{0.32\linewidth}
		\includegraphics[width=\linewidth]{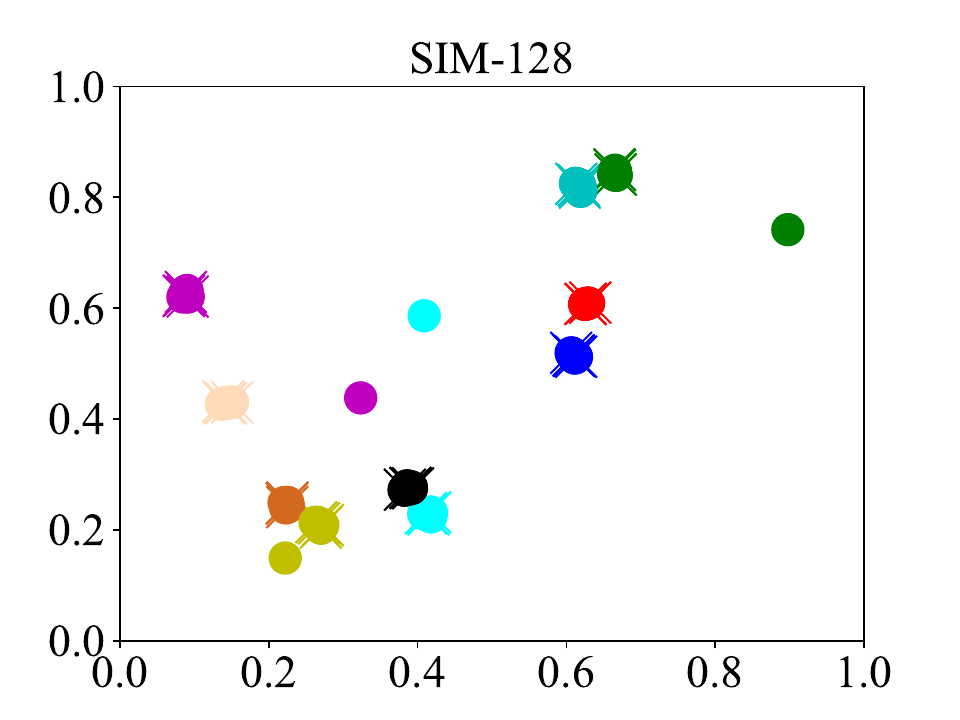}
	\end{minipage}
	\begin{minipage}{0.32\linewidth}
		\includegraphics[width=\linewidth]{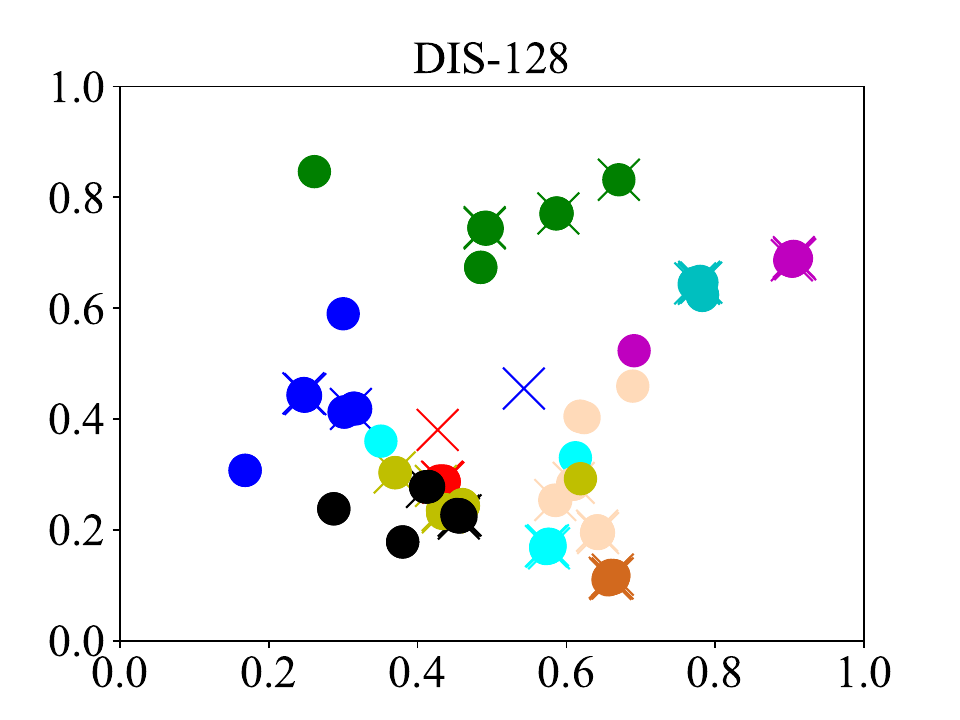} 
	\end{minipage}
	\begin{minipage}{0.32\linewidth}
		\includegraphics[width=\linewidth]{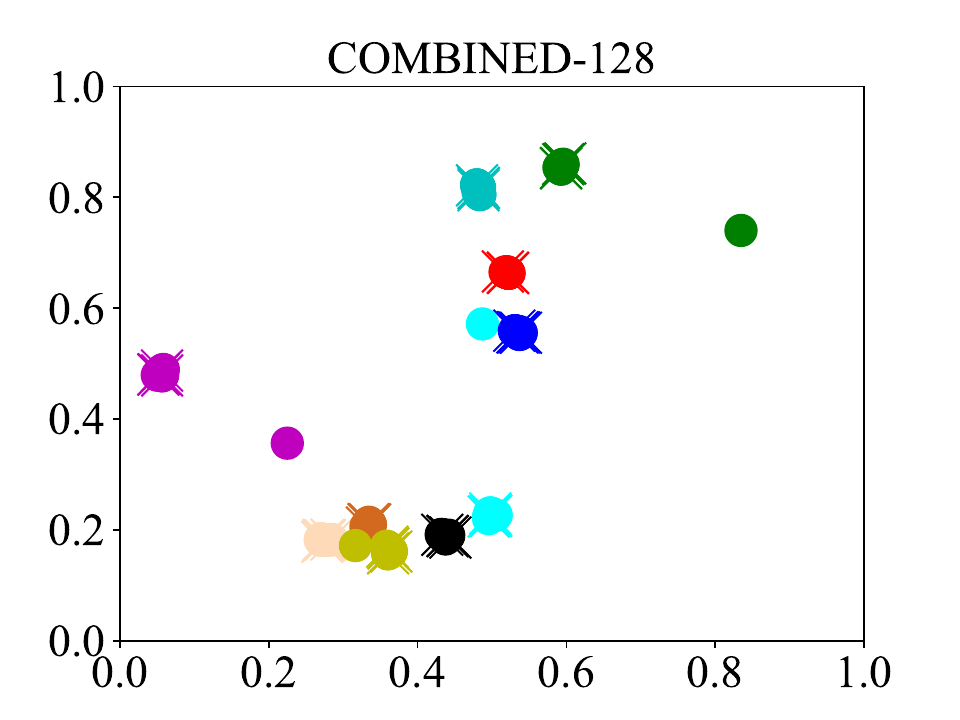}
	\end{minipage}
	\caption{t-SNE visualization of reconstructed features for different metrics.}
	\label{fig:TSNE}
\end{figure*}
In addition to performance evaluation, we analyze the proposed method by examining the reconstructed features. Taking the person ReID task as an example, we visualize the reconstructed features in a 2-dimensional space with t-SNE.
Fig. \ref{fig:TSNE} presents visualizations for 15 randomly selected identities, represented in different colors. The markers ``$\times$'' and ``$\bullet$'' correspond to query images and gallery images, respectively. We visualize reconstructed features under two information capacities, 32 bytes, and 128 bytes, for all three metrics.

Given that person ReID involves searching for the correct gallery image for a query image, we compare the three metrics based on three aspects: 1) whether a query image can find a correct gallery image; 2) the compactness of features extracted from a specific identity; 3) the discriminability between different identities. 
For the information capacity of 32 bytes, more query images are not located near any gallery images, suggesting that more query images fail to find their corresponding gallery images correctly. In addition, the gallery images' compactness for SIM is inferior to the discrimination metrics; for instance, the black dots in SIM are more dispersed than those in the others. It is worth noting that Fig. \ref{fig:TSNE} visualizes only 15 identities, while there are many more identities in reality. The more scattered distribution of one identity, the more easily it can be confused with others. Furthermore, for the five isolated black points, no query images are located near them, implying they are meaningless for identification.
For the information capacity of 128 bytes, both SIM and COMBINED achieve higher compactness and discriminability than DIS. This demonstrates that with sufficient information capacity, a similarity metric can preserve the compactness and discriminability of the original features by maximizing signal fidelity.

\subsection{Ablation on Discriminability of Original Features} 
\begin{figure*}[tbp]
	\centering
	\begin{minipage}{0.4\linewidth}
		\includegraphics[width=\linewidth]{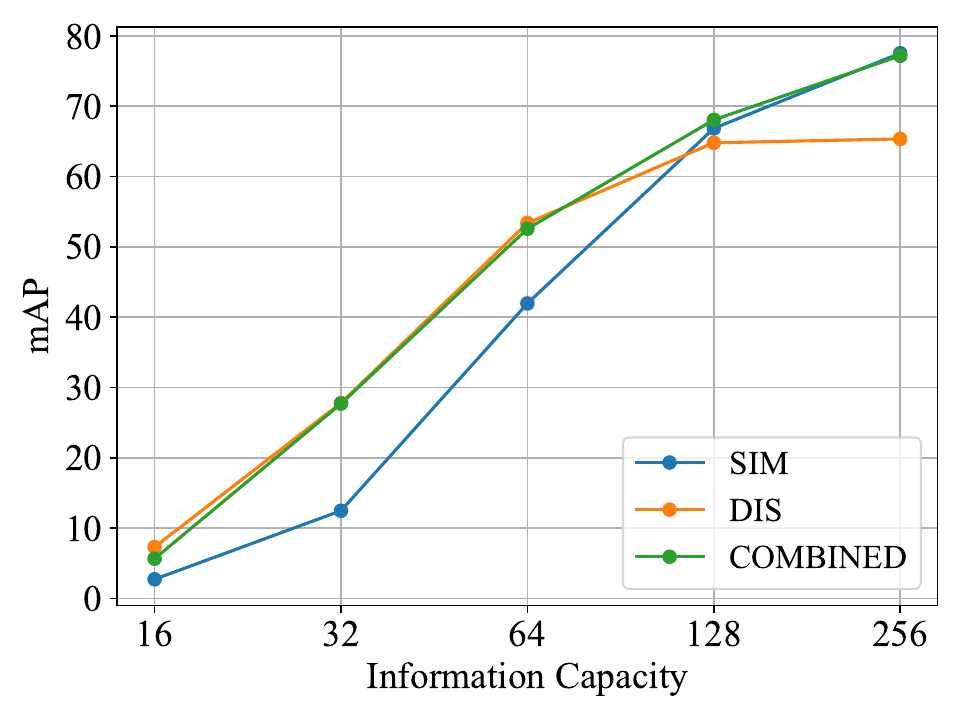}
		\centerline{\footnotesize{(a)}}
	\end{minipage}
	\begin{minipage}{0.4\linewidth}
		\includegraphics[width=\linewidth]{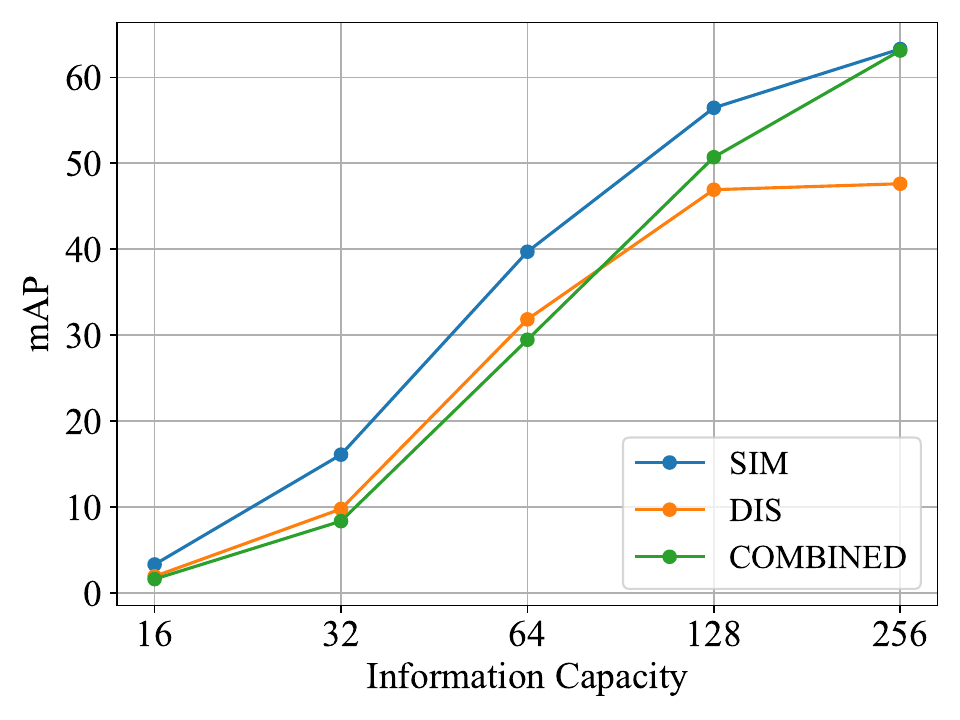}
		\centerline{\footnotesize{(b)}}
	\end{minipage}
	\caption{Evaluation performance comparisons on different metrics. (a) Performance on SimFeat. (b) Performance on DisFeat. }
	\label{fig:ablation_performance}
\end{figure*}
The above experiments verified discrimination metrics help improve the machine vision performance. In this subsection, we delve into the influence of the discriminability of original features on the machine vision performance.
To acquire original features with varying discriminabilities, we train the machine vision network with different metrics. We conduct experiments on the person ReID task.
In previous experiments, the FastReID network was trained with both cross-entropy loss and triplet loss. We denote the features extracted from such a network as SimDisFeat.
In this experiment, we optimize the FastReID network with only cross-entropy loss and only triplet loss, respectively. The features extracted from these two networks are denoted as SimFeat and DisFeat.
In comparison to the default metric, networks optimized for minimizing cross-entropy and triplet loss can produce features with lower and higher discriminability, respectively.
With the trained networks, we extract the original features from the training dataset and then use them to train the compression network. After that, we compress the features extracted from the test dataset and evaluate their performance. The evaluation results are presented in Fig. \ref{fig:ablation_performance}.

\subsubsection{SimFeat } 
In Fig. \ref{fig:ablation_performance} (a), the evaluation performance of three metrics on SimFeat is presented. In terms of relative relationship, discrimination metrics exhibit behavior similar to that observed with SimDisFeat: they outperform at low information capacities while achieving lower performance at high information capacities. However, a difference is notable in the gap between discrimination metrics and MSE. For instance, the gap between DIS and SIM becomes larger, while the gap between DIS and COMBINED becomes smaller. Additionally, DIS achieves comparable performance with COMBINED at the 128-byte information capacity. These differences arise because SimFeat has lower discriminability, indicating that discrimination metrics perform better with features of lower discriminability.

\subsubsection{DisFeat } 
In Fig. \ref{fig:ablation_performance} (b), the evaluation performance of three metrics on DisFeat is presented. Unlike SimDisFeat and SimFeat, DisFeat shows an opposite relative relationship between discrimination metrics and SIM. SIM achieves higher performance at all information capacities. Furthermore, the performance achieved on DisFeat is much lower than that on SimDisFeat and SimFeat. Compared with SimDisFeat, DisFeat has higher discriminability but lower compactness. With discrimination metrics applied, the compression network is optimized to increase the inter-class distance, leading to a further increase in the intra-class distance and lower performance. In contrast, a compression network optimized for SIM can avoid excessively large intra-class distances, resulting in higher performance.

These contrasting experimental results underscore that the trade-off point depends on the discriminability of the original features. The improvement in performance brought by discrimination metrics varies with the discriminability of the original features. The lower the discriminability of the original features, the higher the performance improvement that can be obtained.

\section{Conclusion}
In this paper, we analyzed the feature discriminability in machine vision tasks and addressed its importance in feature compression. Building on this understanding, we introduced a discrimination metric to preserve the discriminability of original features during compression. Through a comparison with the similarity metric, we verified the effectiveness of our proposed discrimination metric. Furthermore, we investigated the trade-off between the discriminability of original features and the proposed discrimination metric.
For future work, we plan to extend the discrimination metric to more machine vision tasks and explore other inter-feature relationships for more feature compression techniques.

\section{Acknowledgement}
This work is supported by the Natural Science Foundation of China under Grant 62036005, and by the Fundamental Research Funds for the Central Universities under Grant WK3490000006.


%\section*{Acknowledgment}

% \newpage


%\bibliographystyle{IEEEtran}
%\begin{thebibliography}{1}
%\end{thebibliography}
\bibliographystyle{IEEEtran}
\bibliography{refs}

\end{document}